# Semantic Properties of Customer Sentiment in Tweets

Eun Hee Ko
Program of Analytics
Northwestern University
Evanston, IL 60208 USA
eunko2013@u.northwestern.edu

Diego Klabjan
Industrial Engineering and Management Science
Northwestern University
Evanston, IL 60208 USA
d-klabjan@northwestern.edu

*Abstract*—An increasing number of people are using online social networking services (SNSs), and a significant amount of information related to experiences in consumption is shared in this new media form. Text mining is an emerging technique for mining useful information from the web. We aim at discovering in particular tweets semantic patterns in consumers' discussions on social media. Specifically, the purposes of this study are twofold: 1) finding similarity and dissimilarity between two sets of textual documents that include consumers' sentiment polarities, two forms of positive vs. negative opinions and 2) driving actual content from the textual data that has a semantic trend. The considered tweets include consumers' opinions on US retail companies (e.g., Amazon, Walmart). Cosine similarity and K-means clustering methods are used to achieve the former goal, and Latent Dirichlet Allocation (LDA), a popular topic modeling algorithm, is used for the latter purpose. This is the first study which discover semantic properties of textual data in consumption context beyond sentiment analysis. In addition to major findings, we apply LDA (Latent Dirichlet Allocations) to the same data and drew latent topics that represent consumers' positive opinions and negative opinions on social media.

## I. INTRODUCTION

Consumers' methods for communicating have been rapidly transforming, and social networking services (SNSs) have become a significant platform where consumers disclose and exchange information about products, retail stores, and e-commerce services. According to recent statistics (Forrester, 2013), the number of social network users will keep growing from 1,732 million today to 2,549 million by 2017. Beyond the simple number, it is understandable why social media is a key topic for practitioners and executives in many business areas when we dive more deeply into the content of users' activities in the new media. A survey by Universal McCann1 in 2009 showed that 32% of the 200 million bloggers worldwide have created posts related to opinions on products and brands and 71% of the 625 million active Internet users view posts produced by bloggers. As identified a new revolution in the relationship with consumers, the ground-breaking communication method used in SNSs has been drawing the interest from business practitioners. This new media "describes a variety of new sources of online information that are created, initiated, circulated and used by consumers on educating each other about products, brands, services, personalities, and issues" (Blackshaw & Nazzaro, 2004).

Text analytics is an emerging technology in business intelligence to derive actionable insights from the web. There are multiple types of sources for text analytics in customer research such as survey comments, call center interactions, and sales notes, but user-generated content on the web has become one of the most popular sources. Firms are competing to use this media to yield business insights in various operations such as customer service and product development. However, even with this tremendous amount of attention on new opportunities, harvesting useful information from textual data on the web is still challenging for information seekers due to the inherent, perplexing features of its form: large volume and structural heterogeneity. Accordingly, research regarding social media and applications of text analytics techniques is still limited in the business context.

This paper aims at finding fundamental characteristics of consumers' opinions enunciated in a textual manner in social media. Specifically, we are interested in detecting semantic similarity or semantic relatedness over a set of documents retrieved from social media and focus on the uniformity and difference between the documents reflecting consumers' opinions, which possibly connote two types of appraisals, positive and negative. In this research, we define document similarity as the distance between terms within documents based on the likeness of their meaning or semantic content. Accordingly, when certain sets of documents show high correlation values, it means that they are semantically similar. We hypothesize that document sets that include consumers' opinions with a sentiment value of the same sign (e.g., positive or negative) would be more homogeneous whereas document sets that include opposite emotional polarities (e.g., documents with a negative view vs. documents with a positive view) would be more semantically independent. We use Twitter as the textual data source for our analysis, because it is one of the most popular SNSs worldwide and the topics on which Twitter users post are not limited. They discuss all types of themes from their everyday lives to political issues and shopping experiences. Therefore, we assume that rich textual data related to shopping experiences are retrieved from Twitter.

We believe that this study is an innovative attempt to discover semantic patterns in a series of users' subjective opinions expressed with a written language on the web, which has not yet been studied especially in consumer study context. In marketing, tweets have been studied but only from the sentiment analysis perspective. In our work, we focus on identifying semantic patterns.

In the following sections, we first discuss our overall approach for this study and then describe the techniques that we adopt, related work, and findings. Finally, we suggest future research directions at the end of this paper.

## II. DATA MANIPULATION

### A. Data Set

The data has been collected over a period of one year based on sets of keywords corresponding to 12 big retailers in the US (e.g., Walmart, BestBuy, Amazon, etc.). The Twitter's streaming API has been used with the keywords matching the names of the retailers, brand names carried at the retailers (e.g., Michael Kors, Fossil, Lenovo) and a few other preselected keywords (e.g., home electronics, geek squad, outdoor rewards).

We used approximately 160,000,000 tweets posted from November 2012 to February 2013, which yields 700,000 to 2,000,000 tweets per day. Original data comprise extraction date and time, user ID, user name, tweets message, and geographical area. Due to the large volume of original data, we divide the data into only with dates, user IDs, and text and conduct further manipulation and analysis based on the three variables. Most of the tweets are written in English, but the raw data set also includes the tweets in other languages such as Chinese or Spanish. We didn't exclude them in data manipulation process since tweets are automatically removed in part-of-speech tagging step, which is programmed to deal with only English words. We, therefore, omit the step extracting English tweets in the initial data manipulation.

### B. Overall Approach

Before we explain each step of the data manipulation process in detail, we must clarify the difference between a document and a set of documents (or a document set). A document is a tweet in this study, and a set of documents is a group of documents extracted with certain criteria or methodologies. For instance, the data set excerpted with the key word "Amazon" is a set of documents that comprises documents (i.e., tweets) with the key word.

Our approach is composed of the following steps: 1) extracting sets of tweets with six key words of six retail store names (e.g., Best Buy, Costco) assuming that the writers of the tweets belong to the consumer group, 2) dividing the six sets of documents retrieved with six key words into two subsets within each document set based on the positive or negative polarities exclusively given to each document (i.e., each tweet) after sentiment calculation is applied to the documents, 3) creating a document consisting of only *adjectives*, *adverbs*, and *verbs* extracted from each subset of documents produced in the previous step, which results in 12 documents from 12 sets of documents (e.g., a document consisting of only *adjectives*, *adverbs*, and *verbs* excerpted from a set of documents with only *negative values* given after sentiment classification is applied to the data set retrieved with the key word *BestBuy*), 4) drawing document similarity and dissimilarity among the documents created in the previous steps, and 5) retrieving topics explaining the two types of documents with positive or negative sentiment polarities. *Figure 1* displays the conceptual diagram of our approach, where the first four steps depict the data manipulation steps with sentiment calculation and part-of-speech tagging and the last two demonstrate real analysis steps, including cosine similarity, k-means clustering, and LDA (Latent Dirichlet Allocation), which are applied to the final data set processed with all of the manipulation methods from the previous steps.

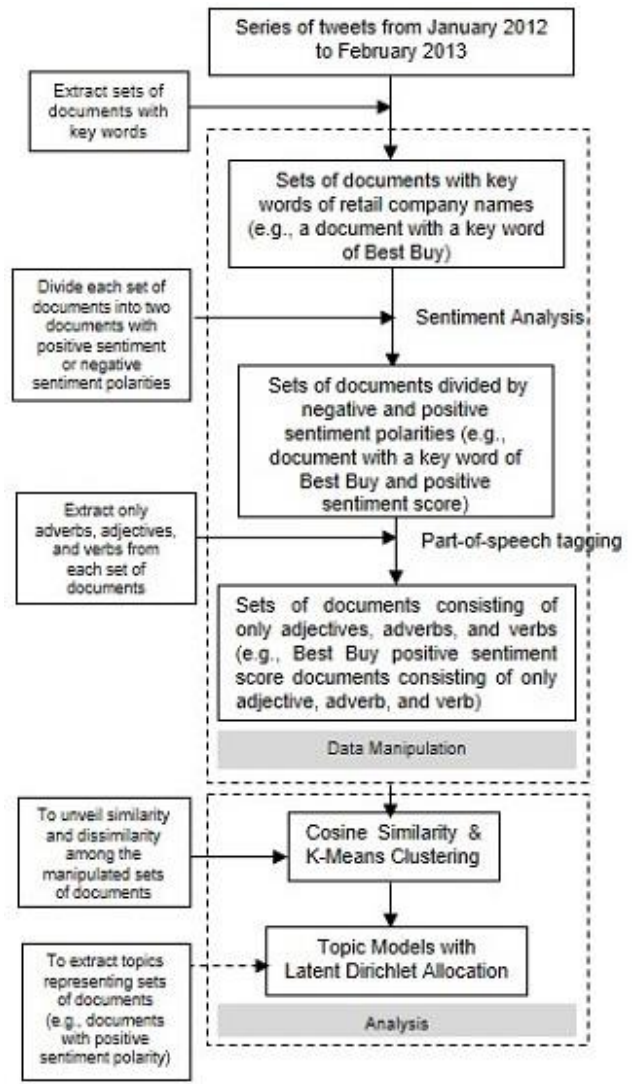

Figure 1. Conceptual diagram of our approach.

### C. Data Extraction with Key Words

Overall, the raw data are processed through three steps as we state in the previous section, Data extraction with key words is the first phase where we retrieve six sets of documents with six retail store names as the key word from the raw data (e.g., Amazon as the key word). Each tweet in the six data sets includes the retail store name corresponding to the key word used to extract the document. *Table 1* shows how many tweets are extracted with each key word from the raw data.

TABLE I. NUMBER OF DOCUMENTS RETREIVED BY KEY WORDS

| Name of document set | Number of document |
|---|---|
| Amazon | 31.526 |
| Bass Pro Shops | 7,195 |
| Best Buy | 35,195 |
| Costco | 28,425 |
| Sam's Club | 21,283 |
| Walmart | 19,993 |

### D. Sentiment Classification

Sentiment analysis is a method for analyzing people's opinions, subjectivities, attitudes, emotions, and sentiments toward entities such as products, services, organizations, issues, events, and topics (Liu, 2012). The English version of an opinion lexicon published by Hu and Liu is adopted to conduct sentiment classification with a simple algorithm that assigns either a positive or negative integer value to each document (i.e., a tweet) by counting the number of negative or positive words in the document.

Before we apply sentiment classification algorithm, we remove stop words, punctuation, and numbers from the data sets and then create six term document matrixes for each data set. We then apply sentiment classification at the document level to further divide the sets of documents extracted with key words into two subsets within each document set: one set with only negative values and the other with only positive values. Each tweet is assigned zero, a negative value, or a positive integer value after the sentiment algorithm is applied. For example, tweets including the key word "Best Buy" have a positive, a negative, or zero sentiment value depending on the words that the documents contain. We discard the tweets with zero value because we are interested only in data analyses with positive or negative sentiment polarities. Finally, we create a total of 12 document sets, two subsets with documents that have either positive or negative sentiment values for the six data sets produced with a key word (e.g., Costco) in the previous step. *Table 2* shows the frequencies of tweets in the 12 sets of documents created through the sentiment classification method.

TABLE II. NUMBER OF DOCUMENTS WITH POSITIVE OR NEGATIVE VALUES

| Name of set of document | Positive | Negative |
|---|---|---|
| Amazon | 5,151 | 1,074 |
| Bass Pro Shops | 2,093 | 635 |
| Best Buy | 3,392 | 2,235 |
| Costco | 5,740 | 2,743 |
| Sam's Club | 5,456 | 3,297 |
| Walmart | 7,322 | 2,721 |

### E. Part-of-Speech Tagging

Another method used in the data manipulation process is part-of-speech (POS) tagging. POS tagging is one form of syntactic analysis that reads text in some language and assigns parts of speech to each word (or each token) such as noun, verb, adjective, etc. *Figure 2* illustrates how POS assigns a tag to each word where "VB" represents verb, "NN" or "NNS" common noun, "JJS" adjective, "IN" preposition, and so forth (Gimpel et al., 2010).

```
"take/VB", "#valentine/JJ", "Best/JJS", "Price/NN"
"Bvlgari/NNP", "Women/NNP", "Eau/NNP", "De/NNP",
Perfume/NNP", "Spray/NNP", "by/IN", "searched/VBN",
"and/CC". "for/IN", "Guide/NNP", "to/TO"
```

Figure 2. Example of part-of-speech tagging.

We apply this technique to produce final data sets consisting of only *adjectives, adverbs,* and *verbs* in each document (i.e., a tweet). We use this data manipulation approach because consumer opinions are usually reflected in these elements, adjective, adverb, and verb, which tend to describe subjects' mobile actions or emotional states as opposed to features (e.g., price) or subjects (e.g., I, Sam), which are usually reflected in nouns or pronouns. The following sentence, which is an actual tweet extracted for this study, displays that adjective, adverb, or verb (the italicized words) represent how this customer feels about "Costco," which possibly demonstrates her/his opinions of this retail chain. However, determining this customer's sentiment based on other elements, such as prepositions or pronouns (e.g., I, all, or also), in this sentence is not possible. The purpose of this study is to find semantic patterns in textual data produced by customer groups. We would like to look at the patterns more clearly, and thus remove other distracting elements in the documents.

```
"I hate all of the evil old farts at Costco.
I also hate that I go to Costco."
```

In detail, we first randomly extract 500 tweets from each of the 12 data sets produced through the sentiment analysis process because we want to have similar size of data over all document sets, but a couple of data sets (e.g., negative set of document for "Bass Pro Shops" in Table 2) have significantly less documents than others. We then apply POS tagging to each set, which consists of the 500 tweets, to extract only adjectives, adverbs, and verbs. This process accordingly produces 12 data sets with 500 documents that include only adjectives, adverbs, and verbs. We call each such data set a *word set*; it consists of the three word elements extracted from the 500 documents. One limitation which we want to point out in the POS tagging process is that we did not perform an evaluation of the tagging performance although it is known that the accuracy of tagging of the OpenNLP POS tagger (i.e., accuracy = *correctly tagged tokens / total tokens*) is usually approximately 0.95. We, however, pre-processed the data by deleting stop words and executed stemming which transfers words to their root form. By standardizing the form of words with the pre-processing treatment, we reduced possible spelling and grammatical errors in the text. In the following sections, the data is consequently composed of 12 word sets. In addition, we will interchangeably use word sets one to six (or seven to twelve) originating from the documents with negative (or positive) sentiment values and *negative (or positive) word sets*.

## III. RELATED WORKS AND FINDINGS

We derive the similarity and dissimilarity between the negative and positive word sets using cosine similarity and k-means clustering techniques. We hypothesize that the word sets with the same sentiment polarity (e.g., positive vs. positive) have higher correlation values than with the word sets with opposite sentiment polarity (e.g., positive vs. negative). As for the k-means clustering method, we anticipate that word sets with the same sentiment sign produce one segment. Before discussing findings, we describe prior works related to this study.

### A. Comparisons with Related Works

A significant number of studies related to text analytics with data retrieved from the web as new forms of online media become more prevalent. In general, there are two streams of academic works related to this research; one is to reinforce algorithms to achieve higher measuring accuracy (Blei, Ng, & Jordan, 2003; Liu, Hu, & Cheng, 2005; Turney, 2002), and the other is to explore new phenomena or practical problems in varied areas by applying the methods to the occurrences (Mishne & Glance, 2006; O'Connor, Routledge, & Smith, 2010). Hu and Liu (2004) studied text summarization of consumer reviews for a product, concentrating on features (e.g., price) of the product with the focus on sentiment analysis. O'Connor, Routledge, and Smith (2010) conduct interesting research which links sentiment analysis for textual data to surveys, which is one of the traditional research methods, and found that correlations of the results from the two methods are high in several cases and seize critical large-scale trends. Our study is more related to the second stream of the academic works, which is an application of text analytics technique to existing phenomenon.

In recent years, business applications of text mining algorithms have surged, and consumer analytics with emerging techniques such as topic modeling (e.g., LSA, LDA), NLP (Natural Language Processing), or sentiment calculation is a big agenda in varied business areas (Ghose & Han, 2011). Major works related to applying text mining to customer analytics are, in particular, committed to finding information about sentiment properties or emotional attitudes that appear in consumer opinions available on the web. Our study differs from previous work in the sense that we explore the opinions of a consumer group beyond the sentiment context by exploiting several methods for the data set from social media. We also try to find fundamental features of consumers' views as well as specific topics that they use in expressing their opinions in the consuming context.

### B. Document Similarity and Dissimilarity

- Cosine Similarity

Cosine similarity is a metric frequently used to discover similarity and dissimilarity among textual types of data. This metric basically calculates the cosine of the angle between two vectors, indicating that cosine 0 degree represents cosines similarity value of 1, which implies that two vectors are exactly the same, and cosine 90 degrees, a cosine similarity value of 0, which means that the vectors are completely independent. Specifically, cosine similarity measures the inner product space between two vectors which are derived from documents. The set of documents is represented as a set of vectors in a vector space where two documents are relatively close in space whenever they are similar in terms of the semantic meaning. For example, vec1 = [1,1,1,1,1,2,1,0,0] and vec2 = [1,1,1,2,0,0,1,1,1] have similarity of 0.9487, which is derived from formula (1). In general, cosine similarity is calculated based on following formula, where *A* and *B* represent two vectors values.

$$\text{similarity} = \cos(\theta) = \frac{A * B}{\|A\|\|B\|} = \frac{\sum_{i=1}^{n} A_i \times B_i}{\sqrt{\sum_{i=1}^{n}(A_i^2)} \times \sqrt{\sum_{i=1}^{n}(B_i^2)}} \quad (1)$$

We generate a term frequency matrix which consists of 12 documents in columns and 1,433 terms in rows with the 12 word sets pre-processed through previous manipulation steps. We apply the cosine similarity method to the term-document matrix to determine similarity and dissimilarity between two forms of word sets: negative word sets from one to six versus positive word sets from seven to twelve. Finally, we conduct calculate the cosine similarity with the term frequency matrix. *Table 3* shows the correlation values. Overall, it reveals that word sets with the same signs are inclined to have higher correlation values.

TABLE III. COSINE SIMILARITY RESULT

| | 1 | 2 | 3 | 4 | 5 | 6 | 7 | 8 | 9 | 10 | 11 | 12 |
|---|---|---|---|---|---|---|---|---|---|---|---|---|
| 1 | 1.00 | | | | | | | | | | | |
| 2 | 0.34 | 1.00 | | | | | | | | | | |
| 3 | 0.27 | 0.08 | 1.00 | | | | | | | | | |
| 4 | 0.52 | 0.23 | 0.31 | 1.00 | | | | | | | | |
| 5 | 0.37 | 0.46 | 0.11 | 0.27 | 1.00 | | | | | | | |
| 6 | 0.44 | 0.54 | 0.16 | 0.37 | 0.69 | 1.00 | | | | | | |
| 7 | 0.57 | 0.05 | 0.02 | 0.10 | 0.06 | 0.05 | 1.00 | | | | | |
| 8 | 0.24 | 0.07 | 0.02 | 0.06 | 0.20 | 0.12 | 0.42 | 1.00 | | | | |
| 9 | 0.45 | 0.02 | 0.02 | 0.08 | 0.08 | 0.05 | 0.81 | 0.48 | 1.00 | | | |
| 10 | 0.50 | 0.01 | 0.02 | 0.09 | 0.07 | 0.04 | 0.73 | 0.46 | 0.91 | 1.00 | | |
| 11 | 0.54 | 0.03 | 0.04 | 0.19 | 0.16 | 0.11 | 0.67 | 0.42 | 0.77 | 0.84 | 1.00 | |
| 12 | 0.44 | 0.05 | 0.06 | 0.14 | 0.29 | 0.16 | 0.60 | 0.39 | 0.72 | 0.67 | 0.65 | 1.00 |

← Negative word sets

Positive word sets ↘

To better display the results, we create the correlation plot (*Figure3*). It demonstrates the correlation trend shown in *Table 3* more clearly, illustrating that word sets two to six (negative word sets) are more strongly correlated and so are word sets seven to twelve (positive word sets). One exception is word set one, a negative word set that is more correlated with positive word sets.

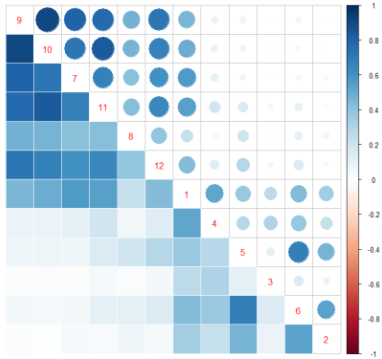

Figure 3. Cosine similarity plot.

- K-Means Clustering

After finding that word sets with the same sentiment polarities tend to be more correlated, we are further interested in how word sets with different sentiment signs are segmented. In detail, we would like to understand how accurately the two kinds of word sets with different sentiment properties are grouped. We use a popular clustering method, k-means. K-means clustering (MacQueen, 1967) is one of the data mining techniques popularly used to split a data set into *k* groups in a way that minimizes the within-cluster sum of squares (WCSS):

$$\arg\min_{\mathbf{S}} \sum_{i=1}^{k} \sum_{\mathbf{x}_j \in S_i} \|\mathbf{x}_j - \boldsymbol{\mu}_i\|^2 \quad (2)$$

where $\boldsymbol{\mu}_i$ is the mean of points in $S_i$ and ($\mathbf{x}_1, \mathbf{x}_2 \ldots \mathbf{x}_n$) is a set of observations.

The data set used in k-means clustering is the same as the one in the cosine similarity study, term frequency matrix. Thus, the k-means method segments the data set based on the frequency of the terms that appear in the term document matrix. We set *k* at two, hypothesizing that there would be two segments and that the word sets originating from the data sets with negative sentiment values (i.e., word sets one to six) would produce one segment, and the word sets from the data sets with positive values (i.e., word sets seven to twelve) would create another segment. *Figure 4* illustrates the k-means analysis result that the positive word sets create one cluster, *cluster1*, and the negative word sets create another cluster, *cluster 2*, while k-means divides the 12 word sets one to twelve into two groups exactly depending on their sentiment polarities. *Table 4* shows that WCSS for cluster 1 is smaller than the one for cluster 2.

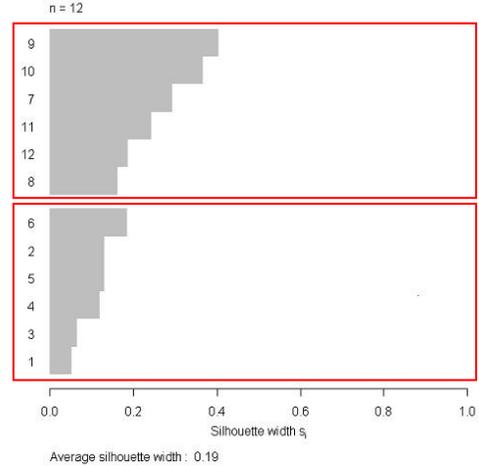

Figure 4. Silhouette plot of K-Means clustering.

TABLE IV. WITHIN CLUSTER SUM OF SQUARES

| Within-cluster sum of squares | Cluster 1 | Cluster 2 |
|---|---|---|
| | 3.42 | 4.18 |
| Between SS/ Total SS | 19.8% | |

C. Topic Model Analysis

Whereas cosine similarity and k-means clustering reveal the overall semantic pattern of the word sets with opposite sentiment polarities, anything related to the content of the documents has not been discovered with these analyses, which is an important business question, "with which terms do consumers describe their opinions about retail stores on social media?" or "what specific semantic properties occur in the posts in which consumers describe their opinions?" To answer these questions, we conduct LDA, which is a technique for topic models, and find the actual content of the text or underlying topics.

- Latent Dirichlet Allocation (LDA)

LDA is a generative probabilistic model for collecting discrete data with a three-level hierarchical Bayesian model, where each item of a collection is modeled as a finite mixture over an underlying set of topics (Blei et al., 2003). This technique is often used in the text modeling context, while the topic probabilities imply an explicit representation of a document (Blei et al., 2003). We apply this technique to discover the underlying topics in the word sets where consumers describe their subjective views of retail stores. The goal of this analysis is to draw overt representations for both word sets including consumer opinion with the positive or negative semantic properties.

Before conducting the LDA, we determine what number of topics best represents each type of the word sets using within groups sum of squares. *Figure 5* shows that three is the optimal number of topics for both positive and negative word sets. We finally perform the LDA with three topics and 10 terms. *Table 5* displays the results.

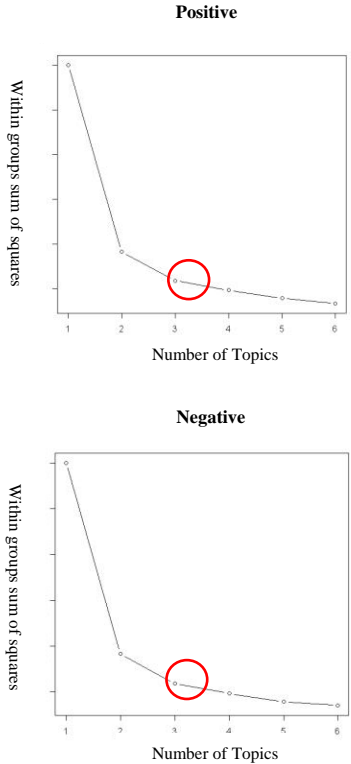

Figure 5. Optimal number of topics.

TABLE V. TOPICS AND TERMS

| Negative Word Sets | | | Positive Word Sets | | |
|---|---|---|---|---|---|
| Topic 1: Controversy | Topic 2: Feature for Price | Topic 3: Controversy | Topic 1: Status | Topic 2: Feature for Newness | Topic 3: Feature for Reward |
| best | big | bad | best | best | free |
| new | bad | crazi | inch | get | love |
| sorri | get | fuck | just | inch | get |
| inch | hate | big | available | just | best |
| black | worst | cheap | led | now | just |
| fuck | crazi | damn | guess | led | now |
| early | fuck | expensive | win | check | good |
| bad | new | first | good | love | win |
| last | cold | old | top | new | got |
| sad | expensive | sad | now | buy | right |

In this study, topic modeling simply reveals what terms embody consumers' positive or negative opinions toward US retail chains in social media. Examining the topics from the LDA analysis, topics 1 and 3 for the negative word sets are labelled as 'Controversy,' because they consist of the words with opposite semantic properties under the same topics. For example, negative topic 1 includes both 'best' and 'bad', which are exactly the opposite semantic meaning. Negative topic 3 also contains both 'expensive' and 'cheap'. We label negative topic 2 as 'Feature for Price' since whereas most of the terms in the topic simply reveal customers' sentiment, the term 'expensive' discovers the reason why customers have negative views on the entities (i.e., retail firms). For the positive word set, we label topic 1 as 'Status', because any term which features the topic does not exist under the topic and most of the terms show customers' sentiment values. We name positive topic 2 as 'Feature for Newness' for the same reason as for negative topic 2; there is term 'new' which characterizes the topic. Finally, we label positive topic 3 as 'Feature for Reward', because it contains such term 'free' which characterizes the topic and most of the other terms simply display customers' emotional status.

Further research should conduct more in-depth level of applications of topic models in marketing context. We suggest that interested researchers collect textual data including consumer opinions in other industries from social media and compare their results of topic models using quantitative methods. It would be also interesting to explore distinguished results of topic models depending on different moderating factors (e.g., demographics of writers).

## IV. CONCLUSION AND FIUTURE RESEARCH

In this paper, we found that the textual data from Twitter reflecting consumer opinions on retail companies show systematic patterns. We hypothesized that documents composed of consumer opinions would be semantically more related to one another when they possess the same semantic polarities. We analyzed term frequency documents that comprise 12 documents in columns (six from the data with negative sentiment value and six from positive) and 1433 terms in rows, using cosine similarity and k-means clustering. Overall, the results confirmed our hypothesis, and showed that documents (i.e., word sets) extracted from data sets with the same sentiment polarity have higher correlation values, whereas the two types of documents with opposite semantic polarities reveal relatively low correlations. In addition, we explored the same data sets with k-means clustering and found that the two forms of word sets produced from tweets exclusively with opposite sentiment assets are accurately separated into two groups: negative documents (i.e., word sets 1-6) comprise one group, and positive documents (i.e., word sets 7-12) create another. After we found the distinct properties of the two types of textual data sets, we conducted LDA to reveal what topics explicitly represent consumers' subjective views on retail companies in social media, which could help business practitioners better understand consumers in emerging media.

We suggest future research that reveals other semantic, fundamental characteristics of consumer opinions not only in social media but also in other communities on the web where users discuss their views of consuming experiences. It would be also interesting to link the sentiment property of consumer opinions on the web to other existing theories in other areas such as psychology or economics.

ACKNOWLEDGMENT

We are much obliged to Matthew Mullins from Aginity LLC (www.aginity.com) to sharing the tweets and providing the computing architecture. Without his help, this work would not have seen the daylight.

ACKNOWLEDGMENT

We are much obliged to Matthew Mullins from Aginity LLC (www.aginity.com) to sharing the tweets and providing the computing architecture. Without his help, this work would not have seen the daylight.